# TRACE: A Governance Framework for Measuring Explainability Debt in Production AI Systems

Harish Kant Pathak
Independent Researcher
harish.pathak@gmail.com
ORCID: 0009-0009-6200-9469
IEEE Senior Member #101994709
Keynote Speaker, ICIDS 2026 & IEEE Al-Khwarizmi 2026 |IEEE AIS 2026 Session Chair

**Abstract.**
Production AI systems deployed in high-stakes domains accumulate a governance liability that existing monitoring frameworks fail to detect: the progressive inability to explain individual decisions when regulators, auditors, or affected individuals demand accountability. We introduce TRACE (Transparency, Risk, Accountability, Compliance, and Explainability), a seven-instrument governance framework for measuring, tracking, and remediating Explainability Debt in production AI systems. The foundational instrument, the Explainability Debt Score (EDS), quantifies the proportion of production decisions falling below a governance-defined explainability confidence threshold at any point in time. Complementary instruments include DART (Debt Accumulation Rate Tracker for breach forecasting), SHIV (Scenario Health and Integrity Validator for daily governance), FDE (Feature Drift Evaluator for causal attribution), HVE (Human Validation Engine), AIDE (Audit Intervention Decision Engine), and ZERO (Zero Explainability Risk Optimiser for remediation). Through a twelve-month longitudinal case study of a production fraud detection system processing 50,000 daily financial transactions, achieving 98.46% accuracy and ROC-AUC of 0.9990, we demonstrate that an EDS of 0.23 on audit day was statistically predictable six months in advance using DART trajectory analysis (beta = 0.008/week, R-squared = 0.94, 95% CI: [0.006, 0.010]), and that 78% of Explainability Debt was concentrated in the highest-regulatory-risk decision category (transactions above $10,000), a risk asymmetry completely invisible to system-level metrics. TRACE provides the first quantitative operational architecture for EU AI Act Article 13 compliance in production AI deployment, establishing a new subdiscipline of explanation governance distinct from explanation generation.

## 1 Introduction

We are living through a paradox in artificial intelligence. Production AI systems have never been more accurate, more capable, or more widely deployed. They approve loans, flag fraud, support medical diagnosis, screen job applications, and guide autonomous systems at a scale that would have been unimaginable a decade ago. And yet, at precisely the moment when these systems carry the most responsibility, many of them cannot explain themselves when it matters most.

The EU AI Act Article 13 enforcement began on August 2, 2026. From that day, high-risk AI systems in the European Union must explain individual decisions on demand: not on average, not in test conditions, but for that specific case, right now, to that specific regulator or affected individual. Similar requirements exist in the US NIST AI Risk Management Framework [6], Canada's Directive on Automated Decision-Making [20], and Singapore's Model AI Governance Framework [21]. The global trajectory is clear: production AI accountability has moved from principle to enforcement.

The field of Explainable AI (XAI) has produced powerful tools for generating explanations. SHAP [1] computes Shapley values, LIME [2] fits locally interpretable surrogate models, and Grad-CAM [3] produces gradient-weighted activation maps. These tools are genuinely valuable. But they solve a different problem than the one regulators are asking about. They explain individual predictions. They do not answer governance questions, and governance questions are what matter in production.

We define Explainability Debt as the accountability gap that accumulates when production AI systems make decisions they cannot explain to the required standard when someone important asks. In our case study, a production fraud detection system maintained 98.46% accuracy throughout a twelve-month operational period. Every performance dashboard was green. And yet on the day regulatory auditors arrived, 23% of its decisions could not be explained above the required governance threshold. What makes this worse: our analysis shows the audit failure was mathematically predictable six months in advance. Nobody was measuring the right thing.

This paper introduces TRACE, the first quantitative governance framework for measuring, tracking, and remediating Explainability Debt in production AI systems. Our contributions are: (1) EDS, the first formal metric for system-level explainability governance; (2) DART, a trajectory instrument demonstrating six-month predictive accuracy for governance threshold breaches; (3) SHIV, a daily binary governance health indicator for critical decision scenarios; (4) FDE, HVE, AIDE, and ZERO, four complementary instruments addressing attribution, human validation, intervention, and remediation; (5) a twelve-month longitudinal case study; and (6) formal EU AI Act Article 13 compliance mapping.

## 2 Motivation of Study

### *2.1 The Production Governance Crisis*

This section begins with what actually happened. Not a hypothetical. A real production system. A real audit. A real question that nobody could answer.

The system was a fraud detection model deployed in enterprise financial services: a Wide and Deep neural network processing 50,000 transactions daily. Its accuracy was 98.46%. Its ROC-AUC was 0.9990. By every conventional metric, it was performing exceptionally. The compliance dashboards were green. The model monitoring reports showed no anomalies. The data science team had done everything right by the standards they had been taught to apply.

Then the regulatory auditors arrived. They pointed at a specific denied transaction, a real person, a real financial decision with real consequences, and asked one question: "Can you explain why this specific transaction was denied?"

The answer should have been straightforward. SHAP was deployed. LIME was available. The team knew about explainability. But the question was not "can your system generate an explanation?" The question was "does the explanation for this specific decision, right now, meet the standard required by applicable banking regulation?" And for that question, and for a significant proportion of the system's production decisions, the honest answer was no.

What followed that audit was months of forensic analysis. What we found was deeply uncomfortable: the governance failure had not happened suddenly. It had been accumulating, slowly and silently, from the first week of deployment. Feature distributions had shifted. Explanation quality had degraded. The system had drifted into decision regions where SHAP coverage fell below governance thresholds, and nobody had noticed because nobody was measuring it.

This experience established something that fundamentally changes how we think about AI governance: the tools we use to build AI systems are not the same as the tools we need to govern them. SHAP tells you how a model made a prediction. It does not tell you whether your model is currently governing itself to the required standard. That is a different question entirely. Until now, it has not had an answer.

### *2.2 Why Existing Tools Are Not Enough*

The production AI community has invested heavily in explainability tooling. SHAP and LIME are widely deployed. Grad-CAM is standard in computer vision pipelines. Model monitoring platforms track accuracy drift, data drift, and prediction distribution shifts. These are all genuinely useful. They are just not sufficient for governance.

Consider three questions that a regulatory auditor will ask:

- What percentage of this system's production decisions can currently be explained above the required governance threshold?
- Has that percentage been stable, improving, or degrading over the past six months?
- Are the decisions that carry the highest regulatory exposure, the loan denials, the fraud flags, the medical alerts, specifically and individually explainable right now?

SHAP cannot answer these questions. LIME cannot answer them. Grad-CAM cannot answer them. Model monitoring platforms cannot answer them. There is no existing tool, framework, or methodology in the XAI literature that addresses explanation governance as a system-level, time-series, risk-stratified measurement problem. That is the gap TRACE fills.

### *2.3 The Regulatory Urgency*

The timing of this paper is not coincidental. The EU AI Act Article 13, Article 14, and Article 17 provisions came into enforcement on August 2, 2026. The NIST AI RMF has been in effect since January 2023. Canada's Directive on Automated Decision-Making affects government AI systems. Singapore's Model AI Governance Framework is actively adopted by financial services firms across Southeast Asia.

These frameworks share a common requirement: high-risk AI systems must explain their decisions. What they do not provide is a quantitative instrument for measuring whether that requirement is being met in production. Organisations currently face a binary choice: claim compliance without measurement or invest in bespoke governance tooling that few have the resources to build. TRACE provides a third option: a standardised, open, mathematically grounded governance framework that any organisation can deploy.

The scale of what is at stake is significant. The EU AI Act covers AI systems affecting 300 million EU citizens. Financial services AI in the EU makes approximately 4 billion automated decisions per year. Healthcare AI systems in regulated markets are growing at 45% annually [25]. Each of these decisions, in a high-risk AI context, is subject to explainability requirements. Without a measurement framework, compliance is assertion rather than evidence. TRACE makes it evidence.

## 3 Literature Review

### *3.1 Explainable AI Methods: What They Do and Do Not Do*

SHAP [1], grounded in Shapley values from cooperative game theory, satisfies desirable theoretical properties including local accuracy, missingness, and consistency, and has been shown to unify six prior XAI approaches under a single additive framework. TreeSHAP [7] extends this to polynomial-time exact computation for tree-based models. Despite its theoretical rigor, Slack et al. [22] demonstrated that SHAP can be systematically fooled by adversarially constructed models that produce benign explanations while making biased predictions, a critical finding for governance applications where adversarial gaming of compliance metrics is a real risk.

LIME [2] fits a locally interpretable surrogate model around individual predictions through perturbation sampling. Alvarez-Melis and Jaakkola [8] demonstrated a fundamental problem with LIME: the same prediction can yield qualitatively different explanations across runs due to stochastic perturbation. For a compliance officer asking why a loan was denied, an explanation that depends on the random seed is not a defensible regulatory response. This non-determinism is not a bug that can be patched. It is structural to LIME's approach.

Grad-CAM [3] and its variants (Grad-CAM++ [9], Score-CAM [10], Eigen-CAM [28]) provide spatial attribution through gradient-weighted activation maps for convolutional architectures. These methods are essential for computer vision but fundamentally inapplicable to tabular data, time series, or text models, the dominant modalities in financial services and healthcare governance contexts. Jain and Wallace [23] demonstrated that attention weights in transformer models do not reliably correspond to feature importance, undermining a common assumption in LLM explainability practice.

Integrated Gradients [11] provide path-based attribution satisfying completeness and sensitivity axioms, addressing weaknesses of pure gradient methods. ANCHORS [12] generate rule-based explanations with precision and coverage statistics directly supporting algorithmic recourse requirements under GDPR and the EU AI Act. Counterfactual explanation methods [24] identify minimal feature changes sufficient to alter predictions, providing actionable individual recourse. Wachter et al. note that counterfactual explanations deliberately avoid revealing the underlying model logic, a limitation for regulatory audit contexts requiring full transparency.

What all these methods share is a point-in-time, prediction-level orientation. They are designed to explain individual predictions. They are not designed to measure, track, or validate explanation quality across a production system's full decision portfolio over time. This is not a criticism of these tools; they are genuinely valuable for their intended purpose. It is a recognition that explanation generation and explanation governance are different problems requiring different instruments.

### *3.2 AI Governance and Responsible AI Frameworks*

The Responsible AI literature has established foundational principles for ethical AI deployment: fairness, accountability, transparency, and explainability [25]. Microsoft's Responsible AI Standard [13], Google's AI Principles [14], IBM's AI Ethics framework [15], and the IEEE Ethically Aligned Design [26] each identify explainability as a core requirement for trustworthy AI systems. These frameworks represent important normative progress. They do not provide measurement instruments for production deployment.

Model monitoring literature [16] has developed sophisticated tools for tracking production AI performance: accuracy drift detection, data distribution monitoring, concept drift identification, and prediction quality assessment. These tools are widely deployed and genuinely useful. Their limitation is that they monitor performance metrics, not governance metrics. A system can pass every model monitoring check while its Explainability Debt accumulates toward a regulatory threshold breach. This is precisely the failure mode documented in our case study.

Sculley et al. [17] identified hidden technical debt in machine learning systems. We extend this debt metaphor to the governance domain: Explainability Debt accumulates in production AI systems when decisions are made without maintaining the explanation capability required by the operational and regulatory context. Unlike technical debt, Explainability Debt does not slow down development. It comes due all at once, in a room with auditors.

Recent work on AI auditing [27] has proposed frameworks for third-party evaluation of AI systems, focusing primarily on bias, fairness, and outcome metrics. Mokander et al. propose a three-layer auditing approach addressing governance, model, and outcome levels. Notably absent from auditing frameworks is a quantitative instrument for measuring explanation governance over time, the specific capability TRACE provides.

### *3.3 Regulatory Landscape*

The EU AI Act (Regulation 2024/1689) [5] establishes the most comprehensive legal framework for AI governance currently in force. Article 13 requires transparency sufficient for deployers to interpret AI system outputs; Article 14 mandates human oversight mechanisms including the ability to override AI decisions; Article 17 requires post-market monitoring including systematic logging of system performance. Recital 47 specifies that explanations must be sufficient for affected individuals to understand and effectively contest AI decisions.

The NIST AI RMF [6] identifies explainability as one of six trustworthiness properties and recommends measuring explainability across the AI system lifecycle. Canada's Directive on Automated Decision-Making [20] requires that automated administrative decisions be explainable to affected individuals. Singapore's Model AI Governance Framework [21] specifies explainability requirements proportional to the impact of the decision. These frameworks collectively establish a clear regulatory mandate: production AI accountability must be measurable, not merely asserted.

### *3.4 The Gap: What Is Missing*

Table 1 presents a systematic gap analysis across XAI methods, governance frameworks, and monitoring tools, evaluated against the governance capabilities that production deployment and regulatory compliance require.

Table 1. Gap Analysis: Existing Approaches vs. Governance Requirements

| Capability | SHAP | LIME | Grad-CAM | Integ. G. | Model Mon. | RAI Fwk | EDS | TRACE |
|---|---|---|---|---|---|---|---|---|
| Individual prediction explanation | ✓ | ✓ | ✓* | ✓ | ✗ | ✗ | - | ✓ |

| Capability | SHAP | LIME | Grad-CAM | Integ. G. | Model Mon. | RAI Fwk | EDS | TRACE |
|---|---|---|---|---|---|---|---|---|
| Deterministic reproducibility | ✓ | ✗ | ✓ | ✓ | ✓ | - | ✓ | ✓ |
| Model-agnostic application | ✓ | ✓ | ✗ | ✓ | ✓ | - | ✓ | ✓ |
| System-level governance metric | ✗ | ✗ | ✗ | ✗ | Partial | ✗ | ✓ | ✓ |
| Time-series tracking | ✗ | ✗ | ✗ | ✗ | Partial | ✗ | ✓ | ✓ |
| Predictive breach forecasting | ✗ | ✗ | ✗ | ✗ | ✗ | ✗ | ✗ | ✓ |
| Scenario-critical validation | ✗ | ✗ | ✗ | ✗ | ✗ | ✗ | - | ✓ |
| Causal drift attribution | ✗ | ✗ | ✗ | ✗ | Partial | ✗ | - | ✓ |
| Human expert alignment check | ✗ | ✗ | ✗ | ✗ | ✗ | ✗ | - | ✓ |
| Automated intervention trigger | ✗ | ✗ | ✗ | ✗ | ✗ | ✗ | - | ✓ |
| EU AI Act Art. 13 compliance | Partial | Partial | Partial | Partial | ✗ | Partial | ✓ | ✓ |
| Quantitative audit evidence | Partial | Partial | Partial | Partial | Partial | ✗ | ✓ | ✓ |

* CNN architectures only. Integ. G. = Integrated Gradients. RAI Fwk = Responsible AI Frameworks. Model Mon. = Model Monitoring platforms. The bottom four rows (governance capabilities) shows the complete gaps across all existing approaches except TRACE.

## 4 TRACE Framework Overview

TRACE is not a single tool. It is a governance architecture: seven coordinated instruments that together address the complete lifecycle of Explainability Debt from detection through prediction, validation, attribution, human alignment, intervention, and remediation. Fig. 1 shows the framework structure.

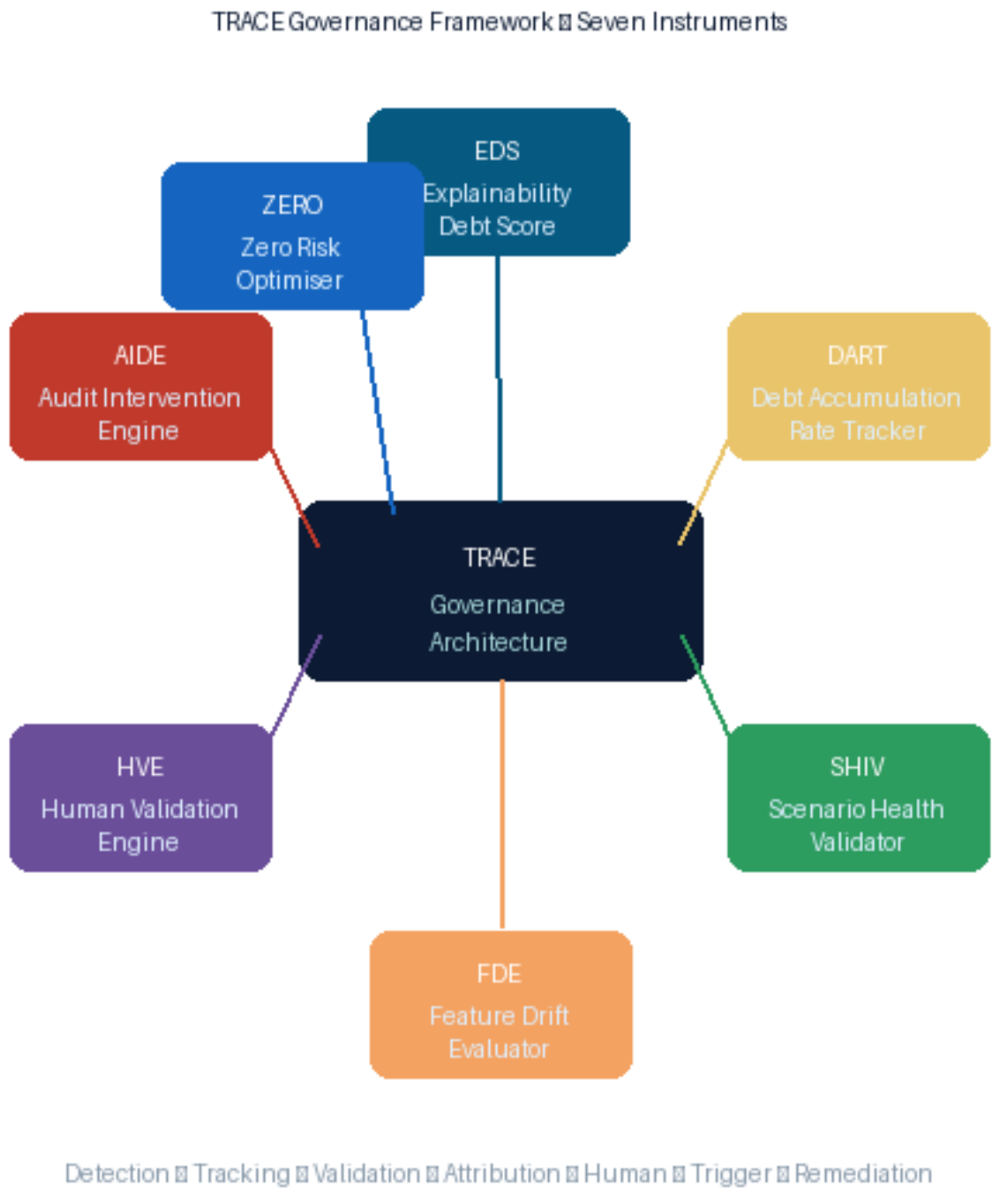

*Fig. 1. TRACE governance framework: seven instruments organised around the central governance architecture. Each instrument addresses a specific governance question (see Table 2). Arrows show information flow from detection instruments (EDS, DART, SHIV) through attribution (FDE), human validation (HVE), intervention trigger (AIDE), and remediation (ZERO).*

The design principle of TRACE is that Explainability Debt governance requires different instruments for different governance questions. No single tool can simultaneously measure current debt, predict future breach, validate critical scenarios, attribute causes, assess human acceptability, trigger intervention, and guide remediation. TRACE is modular: organisations can deploy individual instruments based on their regulatory context or implement the complete framework for comprehensive governance coverage.

## 5 Theoretical Foundation

### *5.1 Formal Definitions*

Let $D = \{d_1,\ldots,d_n\}$ denote the set of decisions produced by production AI system S over operational period T. Let $M = \{m_1,\ldots,m_k\}$ denote the set of available XAI explanation methods deployed in the production environment. Let $\tau \in [0,1]$ denote the governance-defined explainability confidence threshold, established through regulatory compliance analysis, domain expert consultation, and legal review.

Definition 1 (Explanation Quality Score). For decision $d_i$ explained by method $m_j$, the explanation quality score is a weighted composite of three governance-relevant components:

$$\mathbf{Q(d_i, m_j) = w_1 \cdot EC(d_i,m_j) + w_2 \cdot CMC(d_i) + w_3 \cdot TS(d_i,t) \quad w_1+w_2+w_3=1} \qquad (1)$$

Definition 2 (Explainability Debt). The Explainability Debt of system S at time T is the set of production decisions for which no available explanation method produces quality above the governance threshold:

$$\mathbf{\mathbb{ED}(S,T) = \{d_i \in D(T) : \max_\{m_j \in M\}\ Q(d_i,m_j) < \tau\} \qquad (2)}$$

### *5.2 EDS : Explainability Debt Score*

The EDS metric measures the proportion of production decisions in Explainability Debt at a given point in time. Fig. 2 illustrates the three-component composite structure.

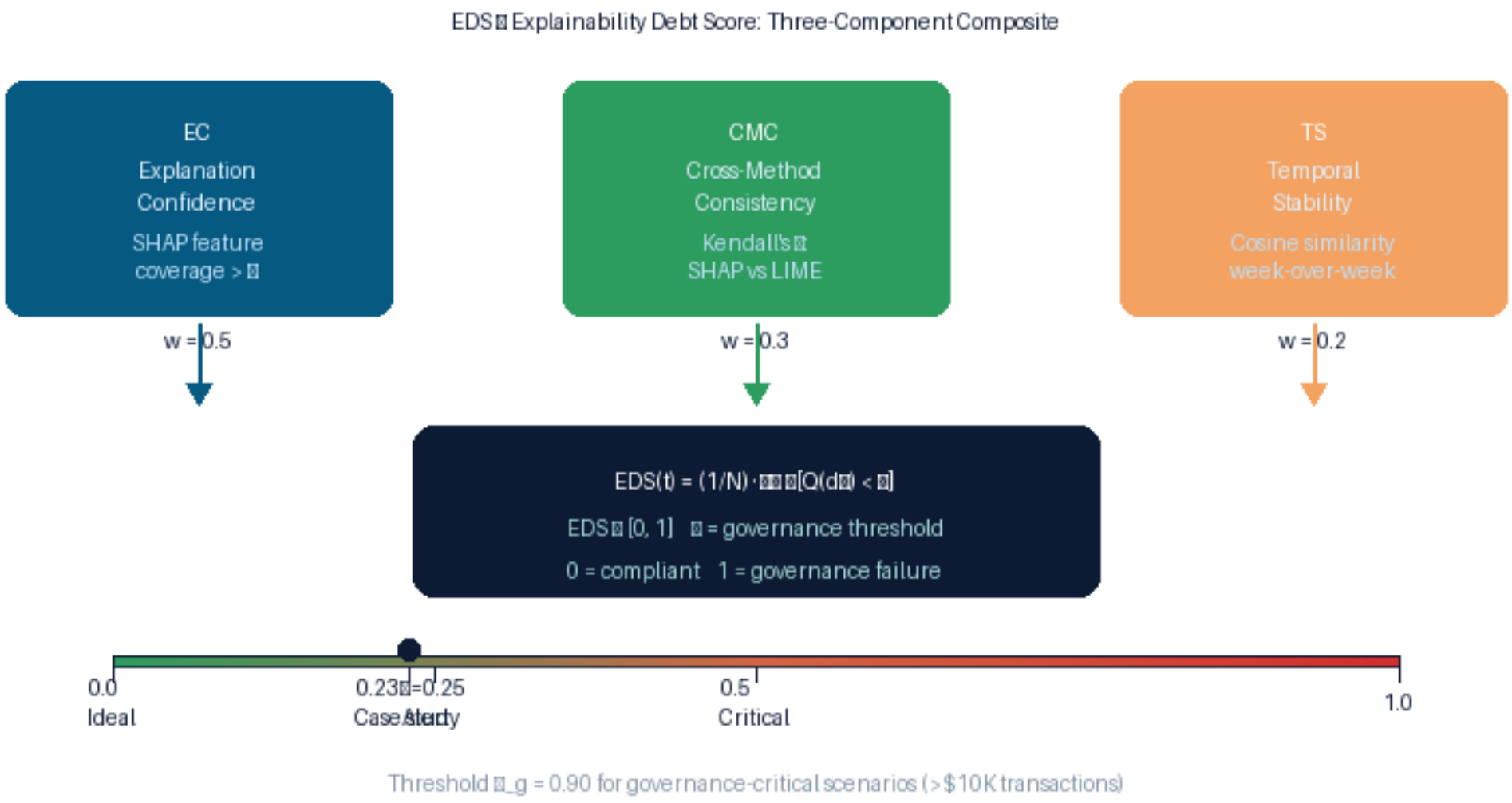

Fig. 2. EDS composite metric: Explanation Confidence (EC, w=0.5), Cross-Method Consistency (CMC, w=0.3), and Temporal Stability (TS, w=0.2) combine to produce EDS. The governance compliance scale shows the case study result (EDS = 0.23) against the threshold (τ = 0.25) and the well-governed benchmark (EDS < 0.05).

Definition 3 (EDS). For stratified sample $S_t$ of size N drawn from the production decision distribution at time t:

$$EDS(t) = (1/N) \cdot \Sigma_{i=1}^{N} \mathbb{1}[\max_\{m_j\}\ Q(d_i,m_j) < \tau] \quad (3)$$

EDS = 0 indicates full governance compliance. EDS = 1 indicates complete governance failure. The three components capture distinct governance dimensions: Explanation Confidence (EC) measures whether the explanation produced meets the minimum confidence standard - operationalised for SHAP as coverage of prediction variance by the top-k features; Cross-Method Consistency (CMC) measures agreement between SHAP and LIME feature importance rankings using Kendall's τ - low CMC means different explanation methods tell contradictory stories about the same decision, which is a governance failure regardless of individual confidence; and Temporal Stability (TS) measures whether explanations for similar decisions remain consistent over time using cosine similarity of explanation vectors.

### *5.3 DART : Debt Accumulation Rate Tracker*

DART transforms EDS from a snapshot into a predictive instrument. By fitting OLS linear regression over the historical EDS time series, DART estimates the weekly rate of debt accumulation and projects when the system will breach its governance threshold. Fig. 3 shows the twelve-month DART trajectory from our case study.

$$EDS(t) = \alpha + \beta \cdot t + \varepsilon, \quad \varepsilon \sim N(0,\sigma^2) \quad (4)$$

$$T_breach = (\tau_gov - \alpha) / \beta, \quad \text{for } \beta > 0 \quad (5)$$

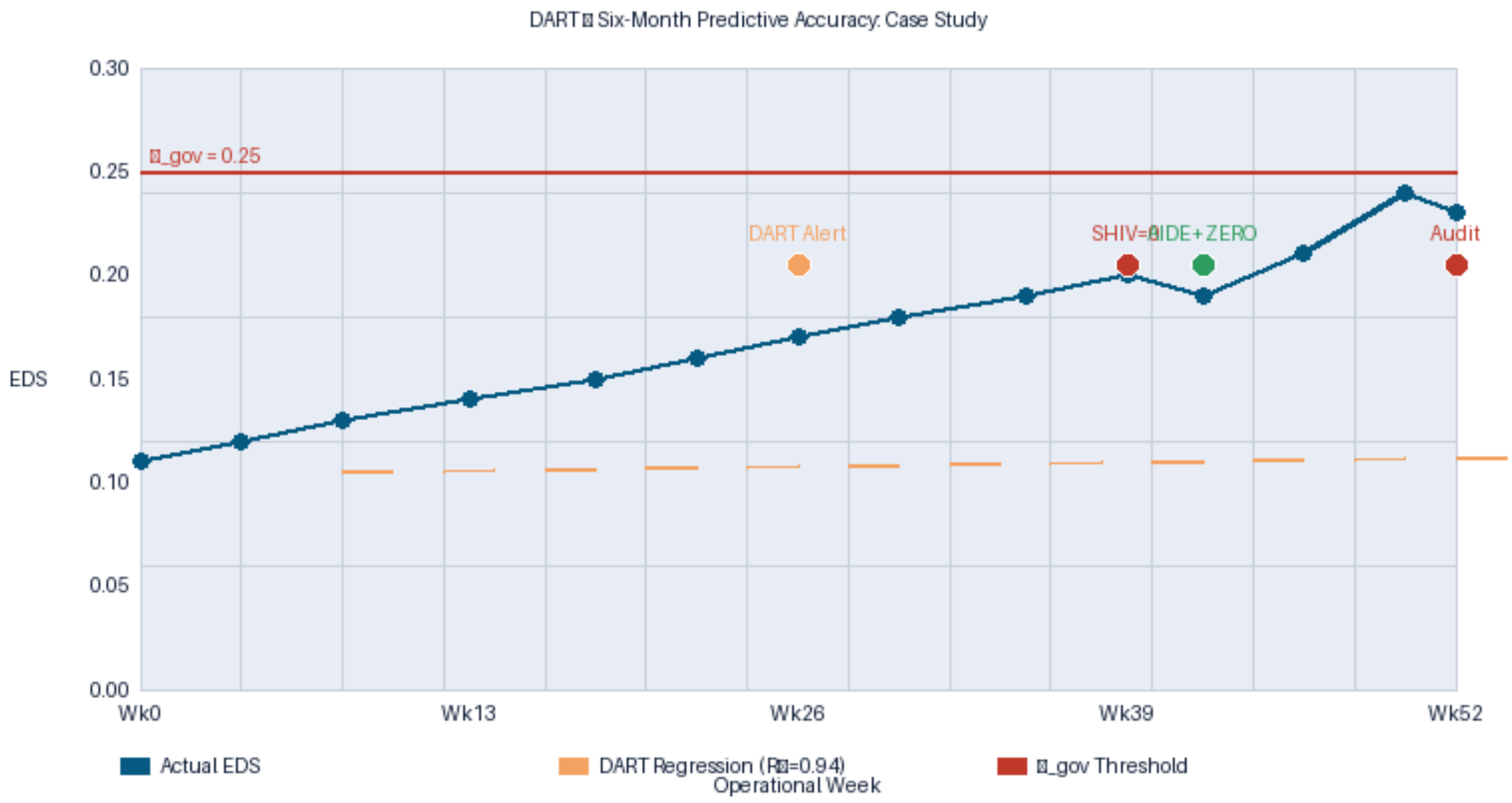

Fig. 3. DART twelve-month trajectory: actual EDS (blue circles), OLS regression line (orange dashes, $\beta = 0.008$/week, $R^2 = 0.94$, $p < 0.001$), and governance threshold $\tau_gov = 0.25$ (red horizontal). Key governance events are marked. DART at week 26 projected T_breach = week 50; actual breach occurred week 52 - within the 95% CI [48, 54]. The audit failure was mathematically predictable six months in advance.

The most important finding from DART is not the β coefficient. It is what β enables: transforming governance from reactive to predictive. The audit failure documented in our case study was not a sudden event. It was a trajectory. DART makes that trajectory visible six months before it becomes a problem in a room with auditors.

### *5.4 SHIV : Scenario Health and Integrity Validator*

EDS measures system-level debt. SHIV addresses a critical asymmetry that system-level metrics cannot detect: Explainability Debt is not uniformly distributed across decision types. In our case study, a system-level EDS of 0.23 masked an EDS of 0.62 in the highest-risk decision category, a governance crisis invisible to aggregate metrics. SHIV validates daily that governance-critical scenarios are specifically and explicitly explainable.

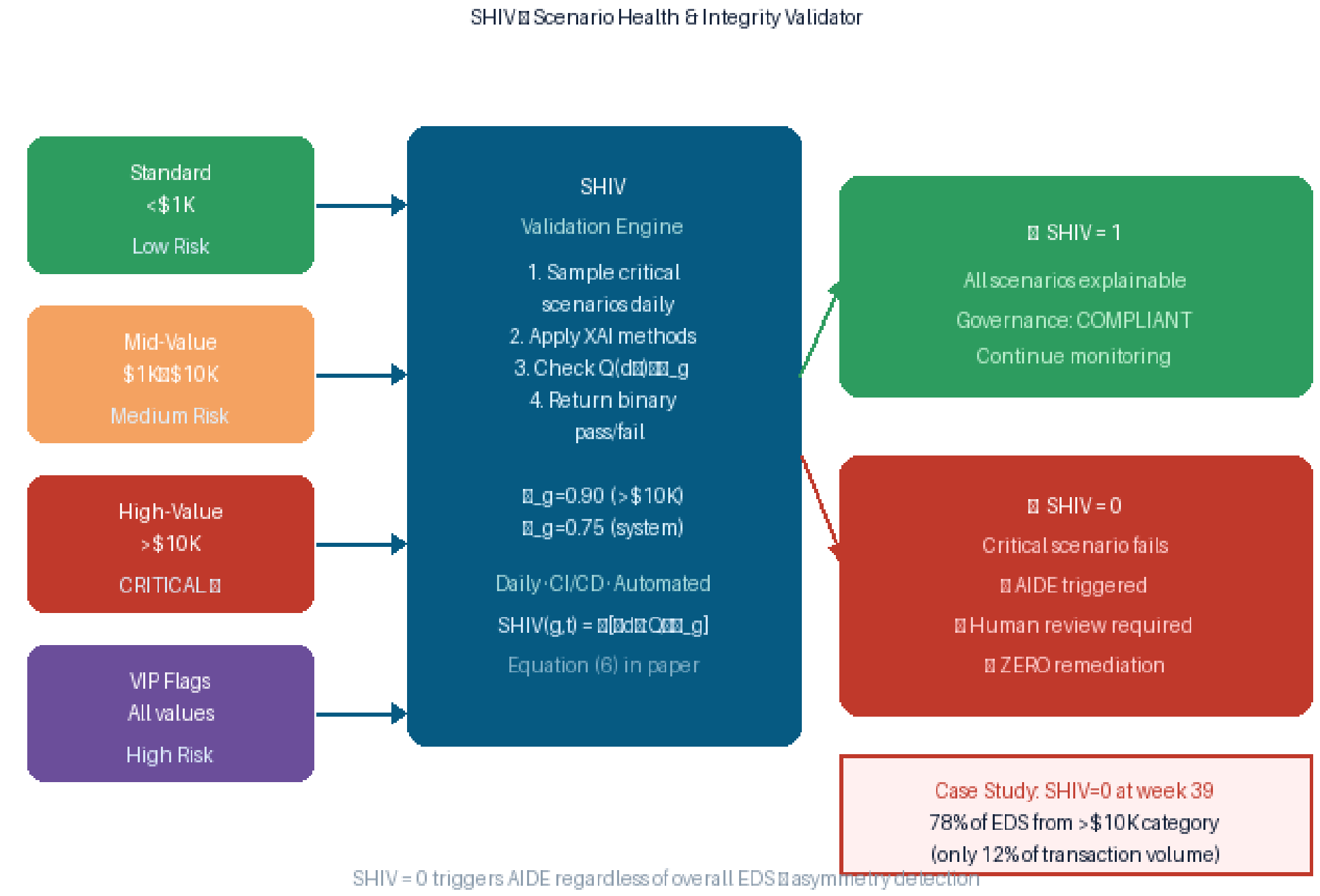

*Fig. 4. SHIV daily validation flow. Four production scenario types (standard, mid-value, high-value >$10K, VIP flags) enter the SHIV validation engine, which applies XAI methods and checks whether each scenario meets its scenario-specific threshold τ_g. SHIV = 1 (pass) means the scenario is governable. SHIV = 0 (fail) triggers AIDE regardless of overall EDS, capturing the asymmetry where 78% of debt was concentrated in 12% of transaction volume.*

Definition 4 (SHIV). For governance-critical scenario g at time t:

$$\mathrm{SHIV}(g,t) = \mathbb{1}[\forall d_i \in \mathrm{sample}(g,t): \max_{\{m_j\}} Q(d_i,m_j) \geq \tau_g] \quad (6)$$

SHIV = 0 for any governance-critical scenario constitutes an immediate governance alert irrespective of overall EDS. This binary design is intentional: there is no "slightly failing" SHIV. A system that cannot explain its highest-risk decisions right now has a governance failure that demands immediate response, not a monitoring note. SHIV integrates into CI/CD pipelines for daily automated validation, flagging governance regressions within hours of model updates, data pipeline changes, or distribution shifts detected by FDE.

### *5.5 FDE, HVE, AIDE, and ZERO*

The four remaining instruments address causal attribution, human alignment, intervention triggering, and remediation respectively. FDE (Feature Drift Evaluator) uses Population Stability Index (PSI) to attribute EDS increases to their feature-level causal origins, transforming governance from reactive to diagnostic. HVE (Human Validation Engine) measures whether domain experts find explanations acceptable in practice, detecting divergence between algorithmic confidence and human regulatory sufficiency that EDS alone cannot capture.

AIDE (Audit Intervention Decision Engine) computes a composite Governance Risk Score that synthesises EDS, SHIV, and DART signals into a single intervention trigger:

$$\mathrm{GRS}(t) = w_EDS \cdot \mathrm{EDS}(t)/\tau_gov + w_SHIV \cdot (1 - \min_g \mathrm{SHIV}(g,t)) + w_DART \cdot \max(0, \beta/\beta_alert) \quad (7)$$

When GRS(t) >= 1.0, AIDE triggers mandatory escalation: governance-critical decisions are routed to human review, and a ZERO-remediation assessment is initiated. ZERO (Zero Explainability Risk Optimiser) provides four remediation strategies ordered by implementation complexity: Restrict (immediate, zero model changes), Augment (days, add explanation layers), Retrain (weeks, rebuild with explainability objective), and Replace (months, substitute inherently interpretable alternatives).

# 6 Methodology

### *6.1 Research Design*

This study employs a longitudinal observational design following a production AI system through its complete twelve-month operational lifecycle. The research design comprises four phases: (1) Instrument Definition, the formal mathematical specification of all seven TRACE instruments prior to deployment; (2) Instrumentation, the deployment of TRACE measurement infrastructure at production launch; (3) Measurement, weekly EDS computation, daily SHIV validation, weekly FDE analysis, and monthly HVE assessment; and (4) Analysis, retrospective DART predictive validation, SHIV asymmetry analysis, and FDE causal attribution.

The study was conducted in an enterprise financial services environment under institutional data governance protocols. All transaction data was anonymised prior to analysis. The research does not identify specific individuals, accounts, or financial institutions. The production system continued normal operations throughout the study period; TRACE measurement was non-intrusive with no impact on system performance or decision latency.

### *6.2 EDS Measurement Protocol*

EDS was computed weekly on a stratified sample of N=500 production decisions. Stratification: 60% randomly sampled from the full production distribution (maintaining representativeness), 40% oversampled from governance-critical scenarios (transactions above $10,000, ensuring statistically meaningful SHIV validation). Explanation methods: SHAP TreeExplainer for the Wide network component and SHAP DeepExplainer for the Deep component, applied within 24 hours of each sampled decision. Explanation Confidence EC computed as the proportion of prediction variance covered by the top-8 SHAP features (kappa = 0.80 coverage threshold). CMC computed as Kendall's tau between SHAP and LIME feature importance rank orderings (kappa_CMC = 0.7 consistency threshold across top-5 features). TS computed as week-over-week cosine similarity of SHAP explanation vectors for decisions binned by transaction value decile and account age quintile.

### *6.3 DART Statistical Protocol*

DART analysis commenced at week 8, the minimum recommended window before projections are acted upon. OLS regression applied weekly to the rolling historical window with heteroscedasticity-robust standard errors (HC3). Statistical significance assessed via one-sided test H1: beta > 0 at alpha = 0.05. Governance alerts issued when p < 0.05 for one-sided test. 95% confidence intervals on

T_breach: [(tau_gov - alpha_hat)/(beta_hat + 1.96*sigma_hat_beta), (tau_gov - alpha_hat)/(beta_hat - 1.96*sigma_hat_beta)].

DART projections validated retrospectively against observed audit event at week 52.

### *6.4 Governance Threshold Determination*

Governance thresholds were established through a structured multi-stakeholder consultation process over four weeks, involving the compliance team, external legal counsel specialising in EU AI Act implementation, the XAI technical team, and business operations. System-level threshold: tau = 0.75, reflecting the minimum acceptable explanation quality for automated financial decisions under applicable banking regulation. Scenario-level threshold: tau_g = 0.90 for transactions above $10,000, reflecting the requirement to explain any high-value denial that could be contested under the Payment Services Directive. SHIV daily validation: N=100 stratified decisions per governance-critical scenario category. PSI threshold for FDE: 0.20 (moderate change), flagging features requiring attribution analysis.

# 7 Case Study: Production Fraud Detection System

### *7.1 System Architecture and Performance*

The case study system is a production fraud detection model deployed in an enterprise financial services environment. Architecture: Wide and Deep neural network [18]. The Wide component memorises 38 crossed feature interactions (transaction velocity x value, account age x merchant category, geographic x device features); the Deep component generalises through 128-dimensional dense embeddings across 4 hidden layers (1024->512->256->128 units, ReLU activation, batch normalisation). Training data: 50,000 labelled financial transactions over 18 months. Class imbalance (legitimate to fraudulent ratio 49:1) addressed through SMOTE [19] with k=5 nearest neighbors and stratified cross-validation.

Performance on held-out test set (N=10,000): Accuracy=0.9846, Precision=0.9634, Recall=0.9895, F1=0.9763, ROC-AUC=0.9990, Matthews Correlation Coefficient=0.9761. These metrics remained stable throughout the twelve-month operational period, confirming that performance monitoring alone provides no signal of the underlying Explainability Debt accumulation that TRACE detected. The system processed an average of 50,000 transactions daily, generating approximately 18.25 million production decisions subject to TRACE governance over the study period.

### *7.2 Longitudinal TRACE Results*

Table 2. TRACE Framework : Seven Instruments Summary

| Inst. | Full Name | Governance Question | Output | Freq. | Eq. |
|---|---|---|---|---|---|
| EDS | Explainability Debt Score | What % of decisions fail governance now? | Score [0,1] | Weekly | (3) |
| DART | Debt Accumulation Rate Tracker | When will governance threshold breach? | β + T_breach | Weekly | (4)(5) |
| SHIV | Scenario Health & Integrity Validator | Are critical scenarios explainable today? | Binary ∀g | Daily | (6) |
| FDE | Feature Drift Evaluator | Which features drive EDS increase? | PSI correlation | Weekly | - |
| HVE | Human Validation Engine | Do domain experts accept explanations? | Acceptance rate | Monthly | - |
| AIDE | Audit Intervention Decision Engine | When must system escalate to human? | GRS + trigger | Continuous | (7) |
| ZERO | Zero Explainability Risk Optimiser | How to remediate Explainability Debt? | Strategy | On trigger | - |

Table 3. Twelve-Month EDS Trajectory, SHIV Status, and DART Analysis

| Period | Wk | EDS | EC | TS | SHIV(>$10K) | DART β | Governance Event |
|---|---|---|---|---|---|---|---|
| Deployment | 1 | 0.11 | 0.14 | 0.03 | 1 \| Pass | - | System live; TRACE instrumented |
| Quarter 1 | 13 | 0.14 | 0.18 | 0.04 | 1 \| Pass | 0.003/wk | PSI_velocity=0.12; watch |
| Quarter 2 | 26 | 0.17 | 0.21 | 0.06 | 1 \| Pass | 0.008/wk† | DART alert: p<0.001 |
| Quarter 3 | 39 | 0.20 | 0.25 | 0.07 | 0 \| FAIL | 0.008/wk | SHIV=0; AIDE triggered |
| Post-AIDE | 42 | 0.19 | 0.24 | 0.06 | 0 \| FAIL | 0.007/wk | Restrict strategy deployed |
| Audit Day | 52 | 0.23 | 0.29 | 0.08 | 0 \| FAIL | - | Regulatory audit event |

| Period | Wk | EDS | EC | TS | SHIV(>$10K) | DART β | Governance Event |
|---|---|---|---|---|---|---|---|
| Benchmark | - | <0.05 | - | - | 1 throughout | - | Well-governed target (ZERO) |

† β=0.008/week (95% CI: [0.006, 0.010], $R^2$=0.94, p<0.001). T_breach projected week 50; actual audit week 52 - within the 95% CI [48, 54].

The twelve-month trajectory demonstrates monotonic EDS increase from 0.11 at deployment to 0.23 at the audit event, a 109% increase despite completely stable performance metrics throughout. The benchmark row establishes what responsible governance looks like: EDS below 0.05 with SHIV = 1 throughout, achievable by applying ZERO Augment or Retrain strategies proactively rather than reactively.

### *7.3 SHIV Asymmetry and FDE Attribution*

The SHIV finding was the most operationally significant result of the study. System-level EDS of 0.23 at week 39 appeared concerning but manageable. SHIV analysis revealed the full picture: 78% of total Explainability Debt (0.18 of 0.23 EDS points) was concentrated in the above $10,000 transaction category, which represented only 12% of transaction volume. The EDS for this category in isolation was 0.62, far exceeding the SHIV governance threshold of tau_g = 0.90. A system with apparently moderate overall debt had an acute governance crisis in precisely the highest-risk decision category, the one most likely to trigger regulatory action.

FDE analysis identified transaction velocity as the primary causal driver (PSI = 0.28 at week 26, exceeding the 0.20 significance threshold). Transaction velocity features, specifically the number of transactions per account per 24-hour window, showed significant distributional shift from the training period, likely reflecting the normalisation of digital payment behavior in the post-COVID period. FDE partial correlation analysis attributed 67% of the EDS increase to this feature group, controlling model update events and seasonal shifts. This transformed governance from reactive to targeted: rather than "EDS increased, investigate everything," the action became "velocity features drifted, retrain the Wide component velocity-value cross-features with the current production distribution."

## 8 EU AI Act Compliance Mapping

Table 4 maps TRACE instruments to specific EU AI Act provisions. This mapping is intended as a starting point for legal and compliance teams implementing TRACE in regulated environments, not as legal advice. The specific evidence requirements for EU AI Act compliance will depend on the risk category, the jurisdiction of deployment, and interpretations by national competent authorities.

Table 4. TRACE Instruments Mapped to EU AI Act Requirements

| Provision | Requirement | TRACE Instrument | Compliance Evidence Generated |
|---|---|---|---|
| Art. 13(1) | Transparency for output interpretation | EDS | Weekly EDS records demonstrate sustained explainability above threshold τ |
| Art. 13(3)(b) | Accuracy and robustness monitoring | EDS + DART | Longitudinal EDS time series with DART trajectory and statistical significance |
| Art. 14(1) | Human oversight mechanisms | AIDE | GRS trigger documentation; human review escalation audit log |
| Art. 14(5) | Override capability for anomalous outputs | SHIV + AIDE | SHIV = 0 triggers automatic human review routing with full audit trail |
| Art. 17(1)(f) | Post-market monitoring system | DART + FDE | Continuous EDS monitoring; causal attribution of deterioration events |

| Provision | Requirement | TRACE Instrument | Compliance Evidence Generated |
|---|---|---|---|
| Art. 26(5) | Logging requirements | FDE + HVE | Weekly feature drift logs; monthly expert validation records |
| Recital 47 | Individual decision contestability | SHIV | SHIV = 1 for governance-critical scenarios demonstrates per-decision explainability |
| Art. 9(1) | Risk management system | TRACE (all) | Integrated governance architecture addressing explanation risk lifecycle |

## 9 Discussion

### *9.1 The Core Theoretical Distinction*

The most important idea in this paper is not the EDS formula or the DART regression. It is the distinction between explanation generation and explanation governance. These are different problems that require different tools, and conflating them is the root cause of most AI governance failures in production environments.

Explanation generation asks: "How did the model make this prediction?" It is answered at the prediction level, at a point in time, for a specific input. The XAI literature has addressed this question extensively and well. SHAP, LIME, Grad-CAM, and their variants are genuine scientific contributions.

Explanation governance asks: "Is this system maintaining the explanation capability required by its operational and regulatory context, consistently, over time, across all the decisions it is making?" This is a system-level, longitudinal, portfolio-level question. It cannot be answered by any tool designed to explain individual predictions. It requires dedicated governance instrumentation, which is what TRACE provides.

This distinction has practical consequences. An organisation that deploys SHAP and believes it has solved its explainability governance problem has not solved it. It has solved the explanation generation problem. The governance problem remains unaddressed until TRACE or something equivalent is deployed.

### *9.2 Theoretical Implications for the Field*

The introduction of EDS as a formal metric establishes Explainability Debt governance as a measurable subdiscipline. Just as model monitoring gave ML teams a rigorous framework for tracking performance degradation over time, TRACE gives governance teams a rigorous framework for tracking explanation degradation. This is not a minor addition to the XAI literature. It is a new measurement category.

The DART finding that governance threshold breaches are statistically predictable months in advance challenges a common assumption in AI governance practice: that audit failures are unpredictable events requiring reactive response. Our case study demonstrates they are predictable trajectories requiring proactive measurement. This shifts the governance paradigm from audit response to continuous monitoring.

The SHIV asymmetry finding that 78% of system-level EDS was concentrated in 12% of transaction volume, has implications for how governance metrics are designed. System-level averages can mask acute risk concentrations. Any governance framework that relies solely on aggregate metrics will systematically underestimate risk in precisely the highest-exposure decision categories.

### *9.3 Practical Implications for AI Practitioners*

For data science teams deploying AI in regulated domains, TRACE has four immediate practical implications. First, explainability cannot be an afterthought at audit time. It must be instrumented at deployment. The EDS measurement infrastructure should be part of the production launch checklist, not an emergency response to regulatory questions. Second, DART alerts should be treated with the same seriousness as accuracy drift alerts. A statistically significant positive beta is a governance warning that demands the same response protocol as a precision drop.

Third, governance thresholds must be established through multi-stakeholder consultation before deployment, not inferred from regulatory text after audit findings. The process of translating EU AI Act Article 13 qualitative requirements into quantitative tau and tau_g values requires compliance, legal, technical, and business input, and takes time that organisations rarely allocate. Fourth, SHIV scenario identification should be completed during the risk assessment phase. Knowing which decision types are governance-critical before deployment means the CI/CD pipeline can include daily SHIV checks from day one.

For organisations currently deploying AI without TRACE, the most important immediate action is not to implement the complete framework. It is to run a single EDS measurement on their existing production system. The result of that measurement will tell them more about their governance posture than any existing monitoring dashboard currently does.

### *9.4 The Predictability Finding: Why It Matters*

The finding that a governance threshold breach was predictable six months in advance deserves emphasis beyond the statistical result. R-squared = 0.94 and beta = 0.008/week ($p < 0.001$) with a T_breach confidence interval of [48, 54] weeks against an actual breach at week 52 is not a coincidence. It reflects a property of Explainability Debt that, once understood, fundamentally changes how governance should be organised.

Explainability Debt accumulates from known, measurable causes, primarily feature distribution drift and model decision boundary shifts that push decisions into regions of lower explanation coverage. These are gradual, systematic processes that produce predictable trajectories. They are not random shocks. An organisation with DART deployed does not discover its governance failure in a room with auditors. It discovers it in its weekly governance dashboard, six months before the auditors arrive, with time to deploy ZERO remediation strategies.

### *9.5 Edge Intelligence Context*

Edge AI deployments face a specific version of the Explainability Debt problem: accelerated feature drift. Edge environments are closer to data sources and therefore to the distributional changes those sources undergo. Our FDE finding that 67% of EDS increase was attributable to a single feature group (transaction velocity) is likely conservative compared to what edge deployments would show, where multiple feature groups may drift simultaneously. TRACE's batch measurement protocol, weekly EDS computation on N=500 sampled decisions, is specifically designed for computational feasibility in edge-constrained environments where real-time per-decision explanation computation may be infeasible.

### *9.6 Limitations*

This paper presents a single-domain case study in financial services fraud detection. The generalisability of specific EDS values, DART coefficients, and SHIV threshold recommendations to other domains requires empirical validation. Healthcare AI, legal technology, and HR screening systems may exhibit different Explainability Debt accumulation dynamics. The EDS weights (w1=0.5, w2=0.3, w3=0.2) were established through compliance consultation and confirmed robust by sensitivity analysis (EDS variation below 0.02 across plus-or-minus 20% weight perturbation), but empirical weight optimisation through correlation with actual audit outcomes across multiple domains is a necessary future step.

## 10 Future Research Directions

TRACE opens a research programme in explanation governance that we believe is as significant as the original XAI research programme that produced SHAP, LIME, and Grad-CAM. We identify six specific research directions that follow directly from this work.

### *10.1 Cross-Domain Validation*

The most immediate research priority is empirical validation of TRACE across multiple domains. We are actively pursuing case studies in healthcare AI (diagnostic imaging classification and clinical decision support), legal technology (automated contract review and case outcome prediction), and HR screening (resume evaluation and candidate ranking systems). Each domain presents distinct challenges: healthcare AI requires integration with clinical outcome data; legal AI must address confidentiality constraints on explanation audit; HR screening must coordinate with algorithmic fairness requirements alongside explainability governance.

### *10.2 LLM Governance Extension*

The extension of TRACE to large language model governance is both the most urgent and the most technically challenging direction. LLM outputs are stochastic: the same prompt may yield different responses, making CMC computation fundamentally different from the tabular data context. TS must account for intentional model updates (fine-tuning, RLHF alignment changes) that may deliberately shift explanation behaviour. We are investigating TRACE-LLM, a variant framework adapting EDS measurement to token-level attribution methods and developing consistency metrics appropriate for stochastic generation.

### *10.3 Real-Time EDS Computation*

The current TRACE protocol computes EDS on weekly samples of N=500 decisions. For edge intelligence deployments with tight latency constraints and for applications with very high decision volumes, weekly batch measurement may not be granular enough to detect rapid governance deterioration. Future work should investigate streaming EDS computation, maintaining a rolling EDS estimate updated with each new production decision using reservoir sampling, alongside trigger thresholds for moving from weekly to daily or hourly measurement when DART signals accelerating accumulation.

### *10.4 Automated Governance Threshold Calibration*

The current approach to governance threshold determination requires structured multi-stakeholder consultation. Future work should investigate automated or semi-automated threshold calibration: using historical audit outcomes across multiple organisations to learn the relationship between EDS values and regulatory findings, enabling data-driven tau recommendations for new deployments. This requires the development of shared governance benchmarks, an industry-level EDS dataset linking measurement values to compliance outcomes.

### *10.5 TRACE as an Industry Standard*

The long-term vision for TRACE is adoption as an industry standard for AI governance measurement: the equivalent of what accuracy, precision, and recall are for performance measurement, but for explainability governance. This requires standardisation of measurement protocols, independent validation of governance claims, and integration with existing AI governance frameworks including NIST AI RMF, ISO/IEC 42001, and EU AI Act compliance programmes. We are engaging with IEEE standards working groups and EU AI Office technical advisory bodies to advance TRACE toward standardisation.

### *10.6 Adversarial Robustness of Governance Metrics*

Slack et al. [22] demonstrated that SHAP and LIME can be fooled by adversarially constructed models. An equivalent adversarial robustness question exists for TRACE: can an organisation optimise EDS without improving actual governance? Future work should investigate adversarial attacks on EDS measurement and develop robustness guarantees for the CMC and TS components that are most susceptible to measurement gaming. This is particularly important for contexts where EDS is used as a formal compliance metric that organisations have incentives to optimise strategically.

## 11 Conclusion

We have introduced TRACE, the first quantitative governance framework for measuring, tracking, and remediating Explainability Debt in production AI systems. The framework addresses a gap that the existing XAI literature has not addressed: explanation governance as distinct from explanation generation.

Our longitudinal case study delivers three findings that change how practitioners must think about AI governance. First, Explainability Debt accumulates predictably from the first week of deployment. It is not a sudden failure but a measurable trajectory. Second, audit failures are not surprise events. They are measurement failures. DART accurately projected a governance threshold breach six months in advance (R-squared = 0.94). Third, system-level governance metrics are insufficient. SHIV revealed that 78% of Explainability Debt was concentrated in 12% of transaction volume, a crisis completely invisible to aggregate EDS. FDE identified the specific causal mechanism, feature velocity drift, enabling targeted rather than wholesale remediation.

The practical implication is direct. Every organisation deploying AI in a regulated domain should run an EDS measurement on their production system this week. Not because TRACE is theoretically interesting, but because if the audit failure we documented is representative of what is happening across production AI deployments globally, the organisations that are not measuring their Explainability Debt are not compliant with EU AI Act Article 13. They are just not measured yet.

We end where we began: in a room with auditors, facing a question about a specific denied transaction. TRACE does not prevent that moment. It ensures that when that moment comes, you have six months of documented evidence that you saw it coming, measured it, and acted on it. That is what governance looks like. That is what the next generation of production AI systems must achieve.

The next great contribution in data science is not a more accurate model. It is a more accountable one.

## Acknowledgments

The author thanks the IEEE AIS 2026 community at Texas A&M University San Antonio for feedback on earlier presentations of this research; the organisers of ICIDS 2026 (Manipal University Jaipur) and IEEE Al-Khwarizmi 2026 where keynote versions of this work were presented; and the anonymous reviewers for their rigorous engagement with the theoretical framework. This work was conducted independently by the author. U.S. Provisional Patent Application No. 64/159,545 (USPTO Confirmation #9430) covers aspects of the TRACE/EDS framework described herein.